\documentclass[12pt,journal,compsoc,onecolumn]{IEEEtran}

\usepackage{times}
\usepackage{epsfig}
\usepackage{graphicx}
\usepackage{amsmath}
\usepackage{amssymb}
\usepackage{pifont}
 \usepackage{slashbox}

\usepackage[utf8]{inputenc} 
\usepackage[T1]{fontenc}    
\usepackage{url}            
\usepackage{booktabs}       
\usepackage{amsfonts}       
\usepackage{nicefrac}       
\usepackage{microtype}      
\usepackage{graphicx}
\usepackage{multirow}

\usepackage{adjustbox}
 
\usepackage[noadjust]{cite}

 \usepackage{color}
 \usepackage{dsfont}
 
\newcommand{\cmark}{\ding{51}}
\newcommand{\xmark}{\ding{55}}

\begin{document}

\title{Learning Chebyshev Basis in Graph Convolutional Networks for Skeleton-based Action Recognition}

\author{Hichem Sahbi \\ CNRS Sorbonne University}

\maketitle

\begin{abstract}

  Spectral graph convolutional networks (GCNs) are particular deep models which aim at extending neural networks to arbitrary irregular domains. The principle of these networks consists in projecting graph signals using the eigen-decomposition of their Laplacians, then achieving filtering in the spectral domain prior to back-project the resulting filtered signals onto the input graph domain. However, the success of these operations is highly dependent on the relevance of the used Laplacians which are mostly handcrafted and this makes GCNs  clearly sub-optimal.\\
In this paper, we introduce a novel spectral GCN that learns not only the usual convolutional parameters but also the Laplacian operators.  The latter are designed "end-to-end" as a part of a recursive Chebyshev decomposition with the particularity of conveying both the differential and the non-differential properties of the learned representations --  with increasing order and discrimination power -- without overparametrizing the trained GCNs. Extensive experiments, conducted on the challenging task of skeleton-based action recognition, show the generalization ability and the outperformance of our proposed Laplacian design w.r.t. different baselines (built upon handcrafted and other learned Laplacians) as well as  the related work. 
\end{abstract}

\section{Introduction}
Deep learning is currently witnessing a major interest in computer vision and different related fields~\cite{ref13,ref17,ref26,Jiu2015,Jiu2016a,Jiu2017,JiuPR2019,icassp2017b,sahbiiccv17}. Its principle consists in training  multi-layered neural networks by designing suitable architectures and optimizing their parameters~\cite{goodfellow2016deep}.  In particular, convolutional networks are well studied and aim at  extracting features that gradually capture {\it low-to-high} semantics of visual patterns. Early   	convolutional networks were dedicated to regular (grid-like) data such as images where convolutions are achieved by shifting equivariant filters and measuring their responses across different image locations. However, data sitting on top of irregular domains (such as skeletons in action recognition) require extending convolutional networks to graph data~\cite{Bruna2013,Defferrard2016,Huangaaa2018,Kipf2016}; while shifting convolutional filters across regular grids  is a straightforward and a well-defined operation, its extension to irregular domains (namely  graphs with heterogeneous topological properties) is  known to be generally {\it ill-posed}. \\ 
Motivated by the success of deep learning in computer vision and machine learning, graph convolutional networks (GCNs)  are currently emerging for different use-cases and applications~\cite{Monti2017}. The common ground of these networks consists in aggregating node representations prior to apply convolutional filters on the resulting node aggregates~\cite{Atwood2016,Gao2018,Niepert2016,Hamilton2017,Monti2017,Zhang2018} (see also \cite{sahbipr2012,sahbicbmi08,Sahbi2015,Sahbi2013icvs,Sahbi2011}). Two categories of GCNs are known in the literature: the first one, dubbed as spatial \cite{Gori2005,Micheli2009,Scarselli2008,Wu2019,Hamilton2017,sahbiicpr2021}, achieves convolution by {\it locally} averaging representations through nodes and their neighbors before applying convolutions using inner products. The second category, known as spectral~\cite{Bruna2013,Defferrard2016,Henaff2015,Kipf2016,Levie2018,Li2018,Zhuang2018,Chen2018,Huang2018,sahbibmvc19}, proceeds differently by first mapping filter and input graph signals using the eigen-decomposition of their Laplacians, then achieving filtering in the resulting spectral domain prior to back-project the filtered signal onto the graph domain~\cite{Slepian1983,Chung97}. While spectral GCNs make convolutions well-defined compared to spatial GCNs, their downside  resides in the non-localized aspect of the learned filters and also in the high complexity of Laplacian eigen-decomposition. \\ 
\indent Other spectral GCNs, known as Chebyshev networks~\cite{Defferrard2016}, consider instead localized convolutional filters using a recursive polynomial decomposition. The success of these particular networks relies on the relevance of the used Laplacian operators which are usually handcrafted or built upon the inherent properties of the targeted applications (e.g., node-to-node relationships in 3D skeletons).  However, handcrafted Laplacians are not able to capture all the relationships between nodes as their setting is agnostic to the targeted tasks. For instance, when considering skeleton-based action recognition, pre-existing node-to-node relationships capture the intrinsic anthropometric aspects of individuals which are necessary for their identification, while other relationships, yet to infer, about their dynamics are necessary in order to recognize their actions (See Fig.~\ref{fig:A2}). Put differently, depending on the task at hand, connectivity in Laplacian operators should be appropriately learned by including not only the available (intrinsic) node-node connections in graphs but also their inferred (extrinsic) relationships.\\
\noindent Current state-of-the art has shifted towards the learning of connectivity in graph signal processing~\cite{Belkin2003,dong18,Daitch,Sardellitti16,Sardellitti19,Valsesia18,Kalofolias,Egilmez,Chepuri,Dong2016} and more recently in GCNs~\cite{Nguyen2019,Miao2020,Jiang2020}. The principle of these methods consists in learning graph connectivity by including explicitly the properties of the underlying Laplacians  during optimization~\cite{LeBars19}. Our proposed method in this paper is different at least in two  aspects; on the one hand, none of this related work considers Laplacian learning as a part of Chebyshev basis design. On the other hand,  existing methods consider multiple independent matrix operators that capture the actual topology of the input graphs and increase the discrimination power of the learned GCN representations, but this comes at the expense of  overparametrized networks and the risk of overfitting. In contrast, our Chebyshev basis design increases the discrimination power of the representations  (that capture different hops in graphs) without overparametrizing the trained networks as the learned Laplacian parameters are shared through all the Chebyshev polynomials. Besides, making the Chebyshev basis\footnote{The generative aspect of our basis makes it possible to capture different hops of neighbors without increasing the actual number of training parameters and this enhances the discrimination power of the learned representation as shown through this paper.} orthogonal allows to control the actual number of training parameters and further enhances the generalization power of our GCNs as corroborated later in experiments.  \\
\indent In this paper, we introduce a novel framework that learns Laplacians as a part of GCN and Chebyshev basis design.  This basis is expressed using an efficient recursive form evaluated on a single shared Laplacian operator which conveys both the differential and non-differential properties of the learned graph representations. This Chebyshev   basis  also  captures the statistical properties of the learned representations, with increasing order  and  discrimination power, without increasing the actual number of training parameters in the resulting GCNs. Different settings  are  considered  in our design including symmetry and orthogonality that further constrain the learned Laplacians and enhance the generalization capacity of our  GCNs. Experiments, conduced on the challenging task of skeleton-based  action and hand-gesture recognition show the high accuracy and the outperformance of our method w.r.t. different baselines as well as the related work.  

 \section{Related work}
 In this section, we discuss the related work both from the methodological and the application point-of-view. This includes graph and Laplacian inference as well as skeleton-based action recognition. \\ 
 
\noindent {\bf Graph and Laplacian inference.} Laplacian inference (or equivalently graph design) is generally ill-posed, NP-hard \cite{Sandeep2019, Hanjun2017,Marcelo2018} and most of  the existing approaches rely on constraints (similarity, smoothness, sparsity, band-limitedness,  etc. \cite{Belkin2003,dong18,Daitch,Sardellitti16,LeBars19,Sardellitti19,Valsesia18,Kalofolias,Egilmez,Chepuri,Dong2016}) for its conditioning \cite{Pasdeloup2017,Thanou2017}. Particularly in GCNs, early methods \cite{Micheli2009,Kipf2016} rely on handcrafted or predetermined node-to-node relationships using similarities or the inherent properties of the targeted applications in order to define Laplacian operators \cite{Kipf2016,YLi2018}. However, in spite of being relatively effective, the potential of these operators  is not fully explored as their design is either agnostic to the tasks at hand or achieved using the tedious cross validation.  More recent  advances aim at defining graph topology that best fits a given task \cite{Yaguang2019,Fetaya2018,Alet2018,Alet2018b,Luca2019,ChenAAAI2020,Zhuang2018,Li2018}. For instance, the work  in \cite{Luca2019} proposes a graph network for semi-supervised classification that learns graph topology with sparse structure given a cloud of points; node-to-node connections are modeled with a joint probability distribution on Bernoulli random variables whose parameters are found using bi-level optimization. A computationally more efficient variant  is introduced in \cite{ChenAAAI2020} using a weighted cosine similarity and edge thresholding. Other solutions make improvement w.r.t. the original GCNs \cite{Kipf2016}  by exploiting symmetric matrices \cite{Li2018} and  discovering hidden structural relations (unspecified in the original graphs), using a so-called residual graph adjacency matrix and by learning a distance function over nodes. The work in \cite{Zhuang2018} introduces a dual architecture with two parallel graph convolutional layers sharing the same parameters. This method considers a normalized adjacency matrix and a positive pointwise mutual information matrix   to capture node co-occurrences through random walks sampled from graphs. \\

\noindent {\bf Skeleton-based action modeling.} Action recognition is one of the main challenging tasks in computer vision \cite{lingsahbi2013,lingsahbieccv2014,lingsahbiicip2014} which has been tackled using RGB-based~\cite{Liu2018}, depth-based~\cite{Wang2018d} and skeleton-based  techniques \cite{Wang2018c}.  In particular, with the emergence of sensors (including Intel RealSense \cite{keselman2017} and Microsoft Kinect \cite{zhang2017}),  interest in  pose estimation and skeleton-based action recognition is increasingly growing \cite{Cao2017}. Early skeleton-based methods are  based on modeling human motions using handcrafted features \cite{Xia2012,Yang2014}, time series and  dynamic time warping~\cite{Vemulapalli2014} as well as   Fourier temporal pyramids  \cite{Wangb2012}. However, most of these techniques are oblivious to the interactions that may exist between the most relevant body parts, i.e., those which are actually involved  in human actions.  Other solutions model these interactions \cite{ref35,ref46,ref57} using skeletal quad \cite{ref8},  Lie group \cite{Vemulapalli2014} and temporal relationships \cite{ref50}. With the resurgence of deep learning \cite{ref13,ref17,ref26}, all these methods have been quickly overtaken  by convolutional and recurrent neural networks \cite{ke2017,ref32,Liu2018,ref36,ref53,Du2015,ref51} as well as  their LSTM variants \cite{ref31,ref36,ref44,Song2017,Zhang2017,Zhang2017b,Lee2017}, and  some of  them rely on attention mechanisms that focus on the most relevant joints in skeletons \cite{Liu2016,ref55}.  With the recent emergence of GCNs  \cite{Shuman2013,Duvenaud2015,Henaff2015}  particularly  in skeleton-based action recognition \cite{Yan2018,Li2019,Huangcc2017,ref20,ref23,Lib2018,Yanc2018,Wen2019}, these models have been increasingly used  for this task as they explicitly model, with a better interpretability,  the spatial and temporal interaction among joints  either separately   \cite{Li2019}  or jointly \cite{Jiang2020}. However, while all  joints contribute in motion, only a few of them are actually relevant to recognize the targeted action categories; hence, other work  focuses on learning more complete spatial and temporal co-occurrences for skeleton data  \cite{Shi2018,Nguyen2019,Zhua2016,Li2018}. \\ 
In all the aforementioned work, none of the existing methods considers the issue of learning connectivity in graphs and Laplacians as a part of Chebyshev GCN  design and this constitutes the main contribution of the following sections.

\def\G{{\cal G}}
\def\V{{\cal V}}
\def\E{{\cal E}}
\def\F{{\cal F}}
\def\S{{\cal S}}
\def\A{{\bf A}}
\def\L{{\bf L}}
\def\I{{\bf I}}
\def\D{{\bf D}}
\def\U{{\bf U}}
\def\U{{\psi({\cal V})}}
\def\VV{{\bf U}}
\def\LL{{\cal L}}
\def\J{{\bf J}}
\def\vec{\textrm{\bf vec}}
\def\tr{{\bf tr}}

\def\Lambdaa{{\Lambda}}
\def\T{{\bf T}}
\def\N{{\cal N}}
\def\W{{\bf W}}

\def\thetaa{{\bf \theta}}
\def\betaa{{\bf \hat{w}}}
\begin{figure*}[hpbt]
\centerline{\scalebox{0.52}{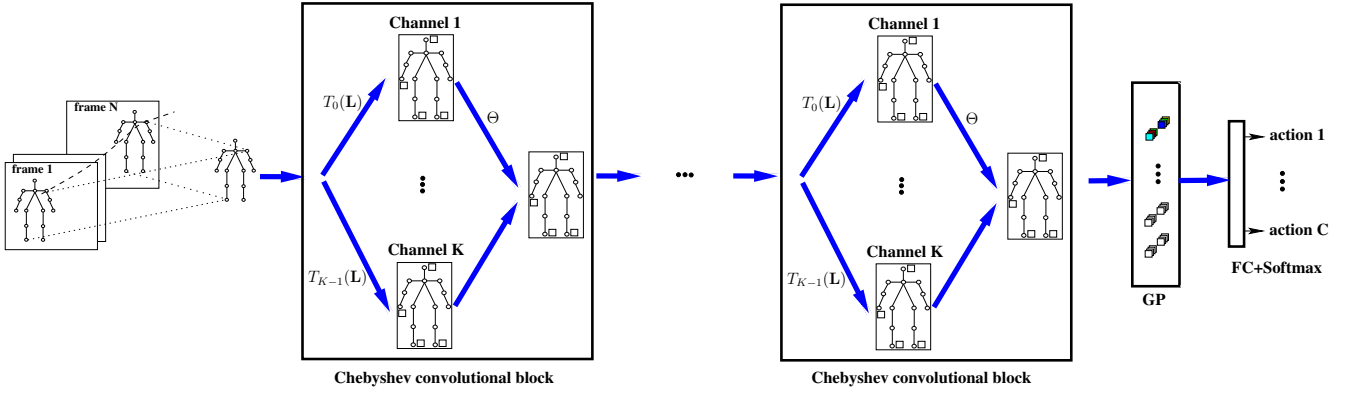}}
\caption{This figure shows the architecture of our Chebyshev Convolutional Network. In each convolutional block, the Chebyshev basis  $\{T_k(.)\}_k $  is first evaluated on the Laplacian $\L$, then multiplied by the input graph signal $\psi(\V)$, and finally aggregated using the parameters in $\Theta$.  These Chebyshev convolutional blocks are followed by global average pooling prior to softmax classification. Note that the Laplacian $\L$ is shared across the basis $\{T_k(.)\}_k $ and through the Chebyshev convolutional blocks. {\bf (Better to zoom the pdf)}.} \label{fig2}
\end{figure*}

\section{Chebyshev Convolutional Networks}\label{cheby}

Let $\S=\{\G_i=(\V_i, \E_i)\}_i$ denote a collection of graphs with $\V_i$, $\E_i$ being respectively the nodes and the edges of $\G_i$. Each graph $\G_i$ (denoted for short as $\G=(\V, \E)$) is endowed with a graph signal $\{\psi(u) \in \mathbb{R}^s: \ u \in \V\}$ and associated with an adjacency matrix $\A$ with each entry  $\A_{uu'}>0$ iff $(u,u') \in \E$ and $0$ otherwise; as shown later in Section~\ref{description},  $\psi(v)$ corresponds to the motion feature of skeletons.  Our  goal is to design a GCN that returns the representation and the classification of a given graph using a novel design of Laplacian convolution on graphs as shown subsequently. \\

\indent Given a graph $\G=(\V,\E)$ with $|\V|=n$, $|\E|$ being respectively the number of its vertices and edges and $\L$ the Laplacian of $\G$; for instance, $\L$ could be the random walk  defined as $\L=\I_n - \A [\D^{-1}(\A)]$  where $\I_n$  is an $n \times n$ identity matrix and $\D(\A)$ a diagonal degree matrix with each diagonal entry $[\D(\A)]_{uu}=\sum_{v} \A_{vu}$. Let $\VV \Lambda \VV^\top$ be the eigen-decomposition of $\L$, with $\VV$, $\Lambda$ being respectively the matrix of eigenvectors (graph Fourier basis) and the diagonal matrix of its eigenvalues; spectral graph convolution is a well defined operator (see for instance \cite{Monti2017})  which is achieved by first projecting a given graph signal $\psi(.)$ using the eigen-decomposition of $\L$, and then multiplying the resulting projection by a convolutional filter prior to back-project the result in the original signal space.  \\
\indent Formally, the convolutional operator $\star_{\G}$ (rewritten for short as $\star$) on a given  graph signal $\psi(\V) \in \mathbb{R}^{s \times n}$  is  $ (\psi \star  g_{\thetaa})_\V =  \VV\  g_{\theta}(\Lambdaa)  \VV^\top  \psi(\V)^\top$;  here $^\top$ is the matrix transpose operator and  $g_{\theta}$ denotes a non-parametric convolutional filter defined as $g_{\thetaa}(\Lambdaa)={diag}(\thetaa)$ with  $\thetaa \in \mathbb{R}^{n}$. As this filter is non-localized, we consider instead \cite{Defferrard2016}        
       \begin{equation}\label{eq00} 
        (\psi \star  g_{\thetaa})_\V = \sum_{k=0}^{K-1} T_{k}(\L)  \ \psi(\V)^\top   \theta_{k},                                                                                                 
        \end{equation}
        \noindent with $\thetaa=(\thetaa_0 \dots \thetaa_{K-1})^\top \in  \mathbb{R}^{K}$ being the learned convolutional filter parameters and $T_{k}$  the $k$-th order Chebyshev polynomial recursively defined as $T_{k}(\L)=2 \L \circ T_{k-1}(\L)-T_{k-2}(\L)$, with $T_{k}(\L) \in  \mathbb{R}^{n\times n}$, $T_{0}(\L)=\I_n$, $T_{1}(\L)=\L$ and $\circ$ the hadamard (element-wise) matrix product. When $\L$ is the combinatorial Laplacian, we consider in practice a rectified version as $2 \L /\lambda_{\textrm{max}} - \I_{n}$  (instead of $\L$ with  $\lambda_{\textrm{max}}$  being the largest eigenvalue of $\L$) in order to guarantee the orthogonality of the basis $\{T_{k}(\L)\}_k$; see again \cite{Defferrard2016} and later (in section~\ref{orthogonality}) the general rectification of any Laplacian that guarantees orthogonality of the Chebyshev basis. \\
 \noindent Using Eq.~\ref{eq00}, the extension of convolution to multiple filters $\F=\{ g_\theta \}_\theta$ and $|\V|$ nodes can be written as 
        \begin{equation}\label{eq01} 
        (\psi \star \F)_\V = \sum_{k=0}^{K-1}  T_{k}(\L)  \U^\top \Theta_k,   
             \end{equation}
here $\Theta=(\Theta_{k})_{k=0}^{K-1}$ is the matrix of convolutional parameters associated to multiple channels (filters). In Eq.~\ref{eq01}, the input signal $\U$ is projected using the Chebyshev  polynomials  $\{T_{k}(\L)\}_k$; in particular, when $k=1$, $T_{k}(\L)=\L$ and this provides for each node $u$, the aggregate set of its neighbors. Taking high order polynomials  capture the $k$ hop neighbor aggregates in $\V$ and makes it possible  to model larger extents and more influencing contexts. When  the Laplacian $\L$ is common to all the graphs and also shared between   $\{T_{k}(.)\}_k$, entries of $\L$ could be handcrafted or learned, so Eq.~\ref{eq01} implements a Chebyshev convolutional block with two layers; the first one aggregates signals in $\N_k(\V)$ by multiplying $\U$ by  $\{T_{k}(\L)\}_k$ while the second layer achieves convolution by multiplying the resulting aggregate signals by the filter  parameters in $\Theta$. The whole architecture of this convolution is described in Fig.~\ref{fig2}.
\section{Our Chebyshev basis design}
The success of the aforementioned convolutional process is highly dependent on the relevance of the Laplacian $\L$  and  knowing a priori which Laplacian (and its hyper-parameters) to
choose could be challenging and usually relies on the tedious cross-validation. One may consider a solution that learns convex combinations of individual Laplacians, each one dedicated to a particular topology of input graph data. However (and as also supported later by experiments), this solution is limited by the modeling capacity of  individual Laplacians; in other words, {\it if none of the individual Laplacians capture the actual topology of the input graphs, then their combinations may also be limited to fully capture this topology}. Our  contribution in this paper aims, rather, at designing convolutional Laplacian operators (``from scratch'') by learning the topological structure of the input graphs. \\
\indent Considering the  tensor of  the Chebyshev polynomials $\{T_{k}(\L)\}_k$ and following Eq~\ref{eq01},  the operations in  $\{T_{k}(\L) \U^\top\}_k$ act as feature extractors that collect different order statistics (including means and variances) of nodes and their  neighbors. For instance, when $\L=\A [\D(\A)]^{-1}$ then  $T_{1}(\L) \U^\top$ models expectations ${\{\mathbb{E}(\psi(\N_k(u)))\}}_u$ and if one considers instead $\L=\I_n- \A [\D(\A)]^{-1}$ then  $T_{1}(\L) \U^\top$ captures, up to a square power, the statistical variance  ${\{\psi(u)-\mathbb{E}(\psi(\N_k(u)))\}}_u$. Hence, $\{T_{k}(\L)\}_k$ corresponds to a basis that extracts different order statistics of graph signals before convolution. \\ 
\noindent Let $\LL$ denote the cross entropy loss associated to a given classification task.  
We turn the design of the Laplacian operator $\L$ (thereby $\{T_{k}(\L)\}_k$) as a part of GCN training;  considering the gradient of $\LL$  w.r.t.  the Chebyshev terms, denoted as $\nabla_k \LL = \frac{\partial \LL}{ \partial T_{k}(\L)}$, and since  $\L$ is shared across $\{T_{k}(\L)\}_k$, one may obtain  

\def\JJ{{\bf J}}

\begin{equation}\label{eq04}
\frac{\partial \LL}{\partial \L} = \vec^{-1}\bigg(\sum_{k=0}^{K-1}   \J_k . \vec(\nabla_k \LL) \bigg),  
   \end{equation}
being  $\J_k \in \mathbb{R}^{n^2 \times n^2}$ the diagonal Jacobian matrix whose entry $[\J_k]_{ij,ij} = \frac{\partial [T_{k}(\L)]_{ij}}{\partial \L_{ij}}$ and $\vec(.)$ a vectorization that appends the entries of a given matrix  using the x-y order in  $\J_k $, and $\vec^{-1}$ its inverse. In the above equation, one may show that  $\frac{\partial [T_{k}(\L)]_{ij}}{\partial \L_{ij}}$ can be recursively obtained as
\begin{equation}\label{eq03} 
   \hspace{-0.3cm} \begin{array}{ll}
   \displaystyle  \left\{ \begin{array}{ll}   0 &  {\footnotesize k=0} \\ 1 &  {\small k =1}   \\ 2 \big[[T_{k-1}(\L)]_{ij} +   \L_{ij}  \frac{\partial [T_{k-1}(\L)]_{ij}}{\partial \L_{ij}} \big] - \frac{\partial [T_{k-2}(\L)]_{ij}}{\partial \L_{ij}}   &  {\small  k \geq 2,}  \end{array}\right.
 \end{array}
                                                                                                                                                                                           \end{equation}
so $\L$ can be updated using Eqs~(\ref{eq04}), (\ref{eq03}) and  stochastic gradient descent (SGD). 
 \begin{table*}
\begin{center}
\resizebox{1.0\linewidth}{!}{
\begin{tabular}{c||c|c}
 &  &     \\
 Constraints                              &      Parametrization  & Jacobian \\
                               \hline
                               \hline 
 
COMB  & $\D({\A^\top})- {\A}$ & $[\JJ_{\textrm{c}}]_{{\small ij,pq}} =  {1}_{\{i=j,p\neq q\}} -  {1}_{\{i\neq j\}}$  \\
 NDRW &     ${\A} . [\D({\A})]^{-1} $  & \hspace{-0.25cm} $[\JJ_\textrm{ndrw}]_{{\small ij,pq}} = {1}_{\{j=q\}}. (\delta_{ip} - \L_{ij})  .[\mathds{1}_n . \D({\A})^{-1}] _{pq}$ \\
 DRW  & $\I_n-{\A} . [\D({\A})]^{-1} $  & $[\JJ_\textrm{drw}]_{{\small ij,pq}} = {1}_{\{j=q\}}. ( \L_{ij}-\delta_{ip}).[\mathds{1}_n . \D({\A})^{-1}] _{pq}$ \\
  NDN  & $[\D({\A^\top})]^{-\frac{1}{2}} {\A} . [\D({\A})]^{-\frac{1}{2}} $  &  \hspace{-0cm} $[\JJ_\textrm{ndn}]_{{\small ij,pq}}={1}_{\{i=p \vee j=q  \}}.   \frac{\L_{ij}}{2\A_{pq}}.\big(2\delta_{ip}\delta_{jq}-[\D(\A^\top)^{-1}\A + \A \D(\A)^{-1}]_{pq}\big)$\\
   DN  & $\I_n-[\D({\A^\top})]^{-\frac{1}{2}} {\A} . [\D({\A})]^{-\frac{1}{2}}$ & \hspace{-0cm} $[\JJ_\textrm{dn}]_{{\small ij,pq}}={1}_{\{i=p \vee j=q  \}}. \frac{\L_{ij}}{2\A_{pq}}.\big([\D(\A^\top)^{-1}\A + \A \D(\A)^{-1}]_{pq}-2\delta_{ip}\delta_{jq}\big)$ \\
\hline 
Symmetry & $\A+\A^\top$ & \hspace{0.2cm} $[\JJ_\textrm{s}]_{{\small ij,pq}} ={1}_{\{(i=p,j=q) \vee  (i=q,j=p)\}}$ \\
\hline 
S-COMB  &    $\D({\A}+\A^\top)- ({\A+\A^\top})$     &\hspace{-1.6cm}  $\JJ_{\textrm{sc}} = \JJ_{\textrm{c}} . \JJ_{\textrm{s}}$ \\
S-NDRW &$(\A+\A') . [\D(\A+\A^\top)]^{-1} $   &\hspace{-1.6cm}  $\JJ_{\textrm{sndrw}} = \JJ_{\textrm{ndrw}}. \JJ_{\textrm{s}}$ \ \\ 
S-DRW &  $\I_n-{(\A+\A^\top)} . [\D(\A+\A^\top)]^{-1} $  &\hspace{-1.6cm}  $\JJ_{\textrm{sdrw}} = \JJ_{\textrm{drw}}. \JJ_{\textrm{s}}$ \\
S-NDN  & $[\D({\A}+\A^\top)]^{-\frac{1}{2}} {(\A+\A^\top)} . [\D({\A+\A^\top})]^{-\frac{1}{2}} $  & \hspace{-1.6cm}  $\JJ_{\textrm{sndn}} = \JJ_{\textrm{ndn}}. \JJ_{\textrm{s}}$ \\ 

S-DN  & $\I_n- [\D({\A+\A^\top})]^{-\frac{1}{2}} {(\A+\A^\top)} . [\D({\A}+\A^\top)]^{-\frac{1}{2}} $  & \hspace{-1.6cm}  $\JJ_{\textrm{sdn}} = \JJ_{\textrm{dn}}. \JJ_{\textrm{s}}$ \\
\hline 
    \end{tabular}}
\end{center} \caption{Different parametrizations and the underlying Jacobians. In this table, COMB stands for ``Combinatorial'' Laplacian, NDRW for ``Non Differential Random Walk'', DRW for ``Differential Random Walk'', NDN for ``Non Differential Normalized'' Laplacian, and DN for ``Differential Normalized'' one. The symmetric variants of these Laplacians are prefixed by "S".} \label{tab000} 
   \end{table*}

\subsection{ Constraining the Laplacian} \label{constrained} 
 As described above, the learned matrix $\L$ is not guaranteed to be a  valid Laplacian\footnote{See for instance \cite{LeBars19} for a comprehensive review of  the properties of valid Laplacian operators.}. In order to further constrain $\L$ to be a valid Laplacian, $\L$ is reparametrized as  $\L=\D(\A)-\A$  which corresponds to the combinatorial form of the Laplacian. If one further constrains $\A$ to be column-stochastic, then $\L$ corresponds to the random walk graph Laplacian which captures the differential properties of graphs; a variant of this operator, dubbed as normalized,  is defined as  $\L=\I_n-[\D(\A^\top)]^{-\frac{1}{2}}\A[\D(\A)]^{-\frac{1}{2}}$.  Note that omitting the left-hand side terms (in the aforementioned Laplacians)  makes it possible to capture the non-differential properties in graphs.  \\
 \indent  With this parametrization of $\L$, one may turn the design of $\L$ into the learning of $\A$ while  guaranteeing   the resulting matrix $\L$ to be  a  valid Laplacian. If one further constrains $\A$ to be symmetric, then all the learned Laplacians will have real eigenvalues and some of them positive semi-definite~\cite{Chung97}; these properties are important when handling indirected graphs and also in Laplacian regularization~\cite{Ando2006}.  Considering these settings of $\L$, the chain rule leads to  

\begin{equation}\label{eq05}
\frac{\partial \LL}{\partial \A} =  \vec^{-1}\bigg(\J . \vec\bigg(\frac{\partial \LL}{\partial \L}\bigg) \bigg),
   \end{equation}
with   $\frac{\partial \LL}{\partial \L}$  obtained from Eq.~\ref{eq04} and $\J$ being a  sparse  Jacobian  matrix whose entry  $[\J]_{ij,pq}=[\frac{\partial \L_{ij}}{\partial \A_{pq}}]_{ij,pq}$; this matrix is given in table~\ref{tab000} for different Laplacian settings including the combinatorial and random walk which   capture the differential and non-differential properties of node features.  We also consider the differential random walk -- as a combination these two Laplacians -- obtained by plugging the latter into the former. All these Laplacians are built upon either symmetric or non-symmetric matrices $\A$.  Note that symmetry is obtained using weight sharing, i.e., by constraining the upper and the lower triangular parts of $\A$ to share the same entries. This is guaranteed by considering a reparametrization as ${\A}+\A^\top$ (with $\A$ being now a free matrix) and by tying  pairwise symmetric entries of the gradient   $\frac{\partial \LL}{\partial \A}$; this is equivalently obtained by multiplying the original gradient $\frac{\partial \LL}{\partial \A}$ by the Jacobian  $[\JJ_\textrm{s}]_{ij,pq}= 1_{\{(i=p,j=q) \vee  (i=q,j=p)\}}$ which is  again extremely sparse and its evaluation is highly efficient. 
\subsection{Orthogonality}\label{orthogonality}  

\noindent   Learning multiple matrices $\{T_k(\L)\}_{k=0}^{K-1}$ allow us to capture different graph topologies when achieving aggregation and convolution, and this enhances the discrimination power of the  GCN representations without increasing the actual number of training parameters (as also shown later in experiments).  However, if aggregation produces,  for a given $u \in \V$,  linearly dependent vectors ${\cal X}_u= \{\sum_{u'} [T_k(\L)]_{uu'}. \psi(u')\}_k$, then convolution will also generate  linearly dependent representations with an overestimated number of training  parameters in the null space of ${\cal X}_u$. Besides, matrices $\{T_1(\L),\dots,T_K(\L)\}$ used for aggregation,  may also correspond to overlapping and redundant neighborhoods. \\ Provided that  $\{\psi(u')\}_{u' \in \N_r(u)}$ are  linearly independent, and $K$  upper-bounded by $\textrm{\bf rank}(\big\{\psi(u')\big\}_{u' \in \N_r(u)})\leq \min(|{\cal V}|,s)$, the condition that makes vectors in ${\cal X}_u$ linearly independent reduces to  orthogonality, i.e.,  $\langle T_k(\L),T_{k'}(\L) \rangle_F=0$,  $\forall k \neq k'$, with $\langle ., .\rangle_F$ being the Hilbert-Schmidt (or Frobenius) inner product defined as  $\langle T_k(\L),T_{k'}(\L)\rangle_F=\tr(T_k(\L)^\top T_{k'}(\L))$ with $\tr(.)$ being  the  matrix trace operator. A sufficient condition that guarantees the orthogonality of  the Chebyshev basis consists in taking the Laplacian  ${2 (\L-\lambda_{\textrm{min}} \I_n)}\slash {(\lambda_{\textrm{max}}-\lambda_{\textrm{min}})}-\I_n$ instead of $\L$ with $\lambda_{\textrm{min}}$ (resp. $\lambda_{\textrm{max}}$) being the smallest (resp. largest) eigenvalue of $\L$, and this guarantees that the eigenvalues of the resulting matrix to be in $[-1,+1]$ and hence the orthogonality (minimality) of $\{T_k(\L)\}_k$ (see for instance \cite{Defferrard2016}). It is easy to see that this normalization equates the rectified Laplacian shown in section~\ref{cheby} (i.e., on the combinatorial setting) as its smallest eigenvalue is  zero.  

\section{Experiments} 
 In this section, we evaluate the performance of our GCN network on the task of action recognition using two challenging skeleton datasets; SBU Interaction~\cite{Yun2012} and First-Person Hand Action (FPHA)~\cite{Garcia2018}. The purpose is to show the relevance of our Laplacian design and its comparison against different handcrafted Laplacians and learned ones as well as more general related work in action recognition. 
\subsection{Datasets and implementation details} \label{description} 
\noindent {\bf Dataset description.} SBU is an interaction dataset acquired (under relatively well controlled conditions) using the Microsoft Kinect sensor; it includes in total 282 moving skeleton sequences (performed by two interacting individuals) belonging to 8 categories:  ``approaching'', ``departing'', ``pushing'', ``kicking'', ``punching'', ``exchanging objects'', ``hugging'', and ``hand shaking''.   Each pair of interacting individuals corresponds to two 15 joint  skeletons and each joint is encoded with a sequence of its 3D coordinates across video frames. In this dataset, we consider the same evaluation protocol as the one suggested in the original dataset release~\cite{Yun2012} (i.e., train-test split). \\ 
The FPHA dataset includes 1175 skeletons  belonging to 45 action categories which are performed by 6 different individuals in 3 scenarios. In contrast to SBU, action categories are highly variable with inter and intra subject variability including style, speed, scale and viewpoint. Each skeleton includes 21 hand joints and each joint is again encoded with a sequence of its 3D coordinates across video frames. We evaluate the performance of our method using the 1:1 setting proposed in~\cite{Garcia2018} with 600 action sequences for training and 575 for testing. In all these experiments, we report the average accuracy over all the classes of actions.\\

\noindent {\bf Skeleton normalization.} Let $S^t=\{p_1^t,\dots, p_n^t\}$ denote the 3D skeleton coordinates at frame $t$. Without a loss of generality, we consider a particular order so that $p_1^t$,  $p_2^t$ and $p_3^t$  correspond to three reference joints (e.g., neck, left shoulder and right shoulder for SBU dataset); as shown in Fig.~\ref{fig:A2}, this corresponds to  joints 2, 4 and 7 for SBU and 1, 3 and 5 for FPHA. As the relative distance between these 3 joints is stable w.r.t. any motion, these 3 joints are used in order to estimate the rigid motion (similarity transformation) for skeleton normalization (see also \cite{Meshry2016}). Each  graph sequence is  processed in order to normalize its 3D coordinates using a similarity transformation; the translation parameters ${\bf t}=(t_x,t_y,t_z)$  of this transformation correspond to the shift that makes the reference point $(p_2^0+p_3^0)/2$ coincide with the origin while the rotation parameters $({\bf \theta}_x,{\bf \theta}_y,{\bf \theta}_z)$ are chosen in order to make the plane formed by  $p_1^0$,  $p_2^0$ and $p_3^0$ coplanar with the x-y plane and the vector $p_2^0-p_3^0$ colinear with the x-axis. Finally, the scaling $\gamma$ of this similarity is chosen to make the $\|p_2^0-p_3^0\|_2$ constant through all the action instances. Hence, each normalized joint  is transformed  as $\hat{p}_i^t= \gamma ({p}_i^t-{\bf t})R_x({\bf \theta}_x) R_y({\bf \theta}_y) R_z({\bf \theta}_z)$ with $R_x$, $R_y$, $R_z$ being rotation matrices along $x$, $y$ and $z$ axis respectively. \\
\begin{figure}[hpbt]
  \begin{center}
    \centerline{\scalebox{0.32}{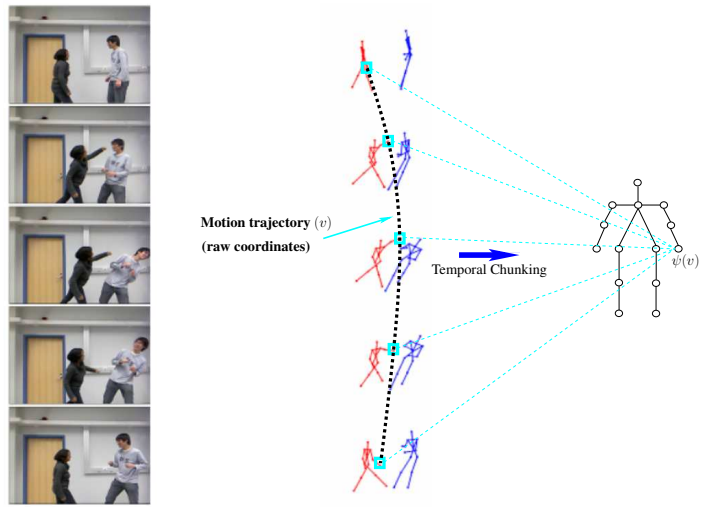}}
\caption{This figure shows the whole keypoint  tracking and description process.} \label{fig1}
\end{center}
\end{figure}

\noindent {\bf Input graphs.}  Considering a sequence of normalized skeletons $\{S^t\}_t$, each joint sequence $\{\hat{p}_j^t\}_t$ in these skeletons defines a labeled trajectory  through successive frames (see Fig.~\ref{fig1}).   Given a finite collection of trajectories,  we consider the input graph $\G = (\V,\E)$ where each node $v_j \in \V$ corresponds to the labeled trajectory $\{\hat{p}_j^t\}_t$  and an edge $(v_j, v_i) \in  \E$ exists between two nodes iff the underlying trajectories are spatially neighbors. Each trajectory (i.e., node in $\G$) is processed using {\it temporal chunking}: first, the total duration of a  sequence (video) is split into $M$ equally-sized temporal chunks ($M=4$ in practice), then the normalized joint  coordinates  $\{\hat{p}_j^t\}_t$  of  the trajectory $v_j$ are assigned to the $M$ chunks (depending on their time stamps) prior to concatenate the averages of these chunks; this produces the description of $v_j$ (again denoted as $\psi(v_j) \in \mathbb{R}^{s}$ with $s=3 \times M$) and $\{\psi(v_j)\}_j$  constitutes the raw description of nodes in a given sequence. Note that two trajectories $v_j$ and $v_i$,  with similar joint coordinates but arranged differently in time, will be considered as very different when using temporal chunking. Note also that beside being compact and discriminant, this temporal chunking gathers advantages --  while discarding  drawbacks -- of two widely used families of techniques mainly {\it global averaging techniques} (invariant but less discriminant)  and  {\it frame resampling techniques} (discriminant but less invariant). Put differently, temporal chunking produces discriminant raw descriptions that preserve the temporal structure of trajectories while being {\it frame-rate} and {\it duration} agnostic.\\

\noindent {\bf Implementation settings.}  We trained the GCN  networks end-to-end using the Adam optimizer \cite{adam2014} for 1,800 epochs  with a batch size equal to $200$ for SBU and $600$ for FPHA, a momentum of $0.9$ and a global learning rate (denoted as $\nu(t)$)  inversely proportional to the speed of change of the cross entropy loss used to train our networks; when this speed increases (resp. decreases),   $\nu(t)$  decreases as $\nu(t) \leftarrow \nu(t-1) \times 0.99$ (resp. increases as $\nu(t) \leftarrow \nu(t-1) \slash 0.99$). All these experiments are run on a GeForce GTX 1070 GPU device (with 8 GB memory) and neither dropout nor data augmentation are used. 

\subsection{Baselines}
We compare the performances of our GCN w.r.t. different Chebyshev basis settings which are defined upon handcrafted and learned Laplacians, as well as against totally learned Laplacian basis. Note that all these Laplacians are combined with symmetry and  orthogonality constraints as described earlier. \\ 

\noindent {\bf Handcrafted Laplacians (HL).} All the Chebyshev terms $\{T_k(.)\}_k$ are evaluated upon a {\it handcrafted} Laplacian $\L$ which in turns depends on a fixed adjacency matrix $\A$ (set using the original input graph). \\

\noindent {\bf Multi-Laplacians (ML).} In this configuration, the Laplacian used in the Chebyshev terms is trained as a weighted {\it  combination} of the handcrafted variants of the Laplacians in table~\ref{tab000} (built upon the fixed matrix $\A$). Note that orthogonality is obtained by normalizing the final learned Laplacian operator while symmetry is enforced in the handcrafted adjacency matrix $\A$. \\

\noindent {\bf Totally Learned Laplacians (TLL).}  In this variant, $K$ independent Laplacians $\{\L_k\}_k$ (and hence the underlying adjacency matrices) are learned. In contrast to the handcrafted setting, orthogonality and symmetry as obtained as a part of the optimization process (as already discussed in sections~\ref{constrained} and~\ref{orthogonality}).

\subsection{Ablation study and comparison}
Tables~\ref{table21},~\ref{table22}  show a comparison of our GCN-based action recognition against the aforementioned GCN baselines, i.e. based on  Handcrafted Laplacians and Totally Learned ones (performances with Multi-Laplacians are rather shown in Table~\ref{compare}); these comparisons are shown for different $K \in \{2,  4, 8\}$.   From all these results, we observe a clear gain of our Chebyshev-based Laplacian design  w.r.t. these baselines; at least one of the setting (namely $K=4$) provides a significant gain.  Table~\ref{table23} shows an ablation study, where the impact of each component  of our GCN (Laplacians, symmetry and orthogonality) is observed separately and jointly. From these results, we  observe a positive impact when constraining the learned matrices to be symmetric and orthogonal; this gain is noticeable with non-differential Laplacians on SBU and with combined (differential/non-differential) ones on FPHA and this clearly shows the complementary aspect of these two Laplacian settings mainly on challenging datasets (i.e., FPHA). Again, this gain reaches the highest values when $K$ is sufficiently (not very) large and this follows the small size of the original skeletons (diameter and dimensionality of the graphs and the signal) used for action recognition which  constrains the required number of Laplacian terms in the Chebyshev decomposition. Hence, with few Chebyshev terms, our method is able to learn relevant Laplacians and representations for action recognition. \\
\indent Our proposed Laplacian design avoids the strong bias about the handcrafted adjacency matrices which are rather suitable to capture the anthropometric characteristics of skeletons and less optimal for action recognition.  On another hand, Tables.~\ref{table21}, \ref{table22} and \ref{table23} show that our Laplacian design makes it possible to capture better the topology of the graph data  (i.e., the neighborhood system defined by the learned Laplacian and its underlying adjacency matrix $\A$). In contrast, the baselines are limited when connectivity is handcrafted and also when learned using totally trained Laplacians, as this results either into a biased Laplacian or into a larger number of training parameters, while Chebyshev provides a compromise between these two extreme cases. Indeed, it enhances the discrimination power of the representation without increasing the actual number of training parameters. In sum, the gain of our GCN results from (i)  the relative flexibility of the proposed design which allows learning complementary aspects of graph topology (through the Chebyshev basis),  and also (ii)  the regularization effect of our constraints (Laplacian weight sharing in Chebyshev, Laplacian parametrization, orthogonality and symmetry) which mitigate overfitting.\\ 
\indent Finally, we compare the classification performances of our GCN against other related methods in action recognition ranging from sequence based such as LSTM and GRU \cite{DeepGRU,Song2017,GCALSTM} to deep graph (non-vectorial) methods \cite{Jiang2020}, etc. (see tables~\ref{compare} and~\ref{compare2} and references within). From the results in these tables,  our GCN brings a noticeable gain w.r.t. related state of the art methods.

 \begin{table}[ht]
 \begin{center}
\resizebox{0.69\columnwidth}{!}{
  \begin{tabular}{cc||c|c|c|c|c}
Laplacians    &  &   \multicolumn{1}{c}{Differential} &   \multicolumn{2}{|c}{Non-Differential}  & \multicolumn{2}{|c}{Combined}  \\ 
 Settings  &  &  \rotatebox{0}{COMB} &   \rotatebox{0}{NDRW} &  \rotatebox{0}{NDN} &  \rotatebox{0}{DRW} &  \rotatebox{0}{DN} \\ 

 \hline
  \hline
     \multirow{3}{*}{\rotatebox{0}{$K=2$}}  & HL    &     96.9231 &    96.9231   &  96.9230   &  93.8462 &  96.9230  \\  
                                            & TLL&   96.9231  &    95.3846  &    98.4615  &  96.9231  &   98.4615   \\ 
                                            &  Our  &   98.4615 &   98.4615   &   96.9230 & 96.9231   &  98.4615  \\ 

  \hline
  \multirow{3}{*}{\rotatebox{0}{$K=4$}}     &  HL &  95.3846  &    93.8462  &  96.9231 &  96.9230  & 96.9230  \\ 
                                             &TLL &    96.9231 &   95.3846  &   98.4615  & 98.4615   &  96.9230 \\ 
                                             & Our &   98.4615   &    \bf100.000  &    98.4615 & 98.4615   &  98.4615  \\ 

 \hline 
  \multirow{3}{*}{\rotatebox{0}{$K=8$}}     & HL &  96.9231 &  98.4615  &   96.9231   &96.9230    & 96.9230  \\ 
                                             & TLL &    96.9231   &      98.4615    &       98.4615 & 93.8462 &  96.9230 \\ 
                                             & Our &    96.9231  &    98.4615  &     98.4615 &  98.4615  & 98.4615  \\ 

 \hline 
    \end{tabular}}
\end{center}
\caption{Detailed performances on SBU using Chebyshev networks with handcrafted (HL) and learned Laplacians (Our), and using totally learned Laplacians (TLL). These performances are shown for $K \in \{2,4,8\}$ and for different parametrizations of the Laplacians including differential (COMB), and non differential (NDRW, NDN) as well as their combinations (DRW, DN); see again Table.~\ref{tab000}. Note that both symmetry and orthogonality constraints are used in these results.}\label{table21}

\end{table}
 \begin{table}[ht]
 \begin{center}
\resizebox{0.69\columnwidth}{!}{
\begin{tabular}{cc||c|c|c|c|c}
Laplacians    &  &   \multicolumn{1}{c}{Differential} &   \multicolumn{2}{|c}{Non-Differential}  & \multicolumn{2}{|c}{Combined}  \\ 
 
Settings  &  &  \rotatebox{0}{COMB} &   \rotatebox{0}{NDRW} &  \rotatebox{0}{NDN} &  \rotatebox{0}{DRW} &  \rotatebox{0}{DN} \\   
 \hline
  \hline
     \multirow{3}{*}{\rotatebox{0}{$K=2$}}     & HL  & 85.9130 & 85.3913 & 85.5652    & 85.3913      & 84.8695  \\   
                                                & TLL & 85.5652 & 86.4348 &  85.7391 & 85.3913  & 86.0869  \\
                                                & Our & 85.3913 & 85.7391 & 85.5652    &  85.5652   & 85.7391   \\

  \hline
  \multirow{3}{*}{\rotatebox{0}{$K=4$}}        & HL  & 86.4348 & 84.1739 & 85.9130 & 84.0000        & 84.5217   \\
                                                & TLL & 84.3478 & 85.3913 &  86.4347   & 85.0435      & 85.5652  \\
                                                & Our & 85.2174 & 85.3913 & 85.7391 & \bf87.1304     & \bf87.3043   \\

 \hline 
  \multirow{3}{*}{\rotatebox{0}{$K=8$}}        & HL  & 85.2174 & 83.8261 & 86.0869   & 84.6957       &85.7391 \\
                                                & TLL & 84.5217 & 85.5652 & 85.7391 & 85.0435         &84.8695 \\
                                                & Our & 84.6957 & 86.9565 & 86.7826  &   84.6957    & 84.5217 \\

 \hline 
    \end{tabular}}
\end{center}
\caption{Same caption as Table.~\ref{table21} on the FPHA database.}\label{table22}
\end{table}

 \begin{table}[ht]
 \begin{center}
\resizebox{0.69\columnwidth}{!}{
  \begin{tabular}{cc||cc||c|c|c|c|c||c}

    &  & \multicolumn{2}{c||}{Constraints} &  \multicolumn{1}{c}{Differential} &   \multicolumn{2}{|c}{Non-Differential}  & \multicolumn{2}{|c||}{Combined} & Avg. \\

Dataset  &  &  \rotatebox{0}{Sym} &  \rotatebox{0}{Orth} &   \rotatebox{0}{COMB} &   \rotatebox{0}{NDRW} &  \rotatebox{0}{DRW} &  \rotatebox{0}{NDN} &  \rotatebox{0}{DN} & perf.  \\ 
 \hline
  \hline
  \multirow{4}{*}{\rotatebox{0}{SBU}} &  &   \xmark &  \xmark &      90.76    & 98.46    & 98.46 & 96.92  & 98.46   &  96.61
 \\ 
                                      &  & \cmark & \xmark &   95.38      &  100.0       & 98.46    & 98.46   & 98.46   & 98.15 \\ 
                                      &  & \xmark  & \cmark &   95.38      & 96.92       & 98.46    & 96.92    &   98.46    &  97.23 \\
                                      &  & \cmark  & \cmark &  98.46      &   \bf100.00   & 98.46    & 98.46   & 98.46   & \bf 98.76 \\
               Avg.                &  &    -     &  -     &   95.00          & \bf 98.84     &      98.46            &     97.69    &  98.46           &  -\\ 
  \hline
  \hline
  \multirow{4}{*}{\rotatebox{0}{FPHA}}     &  &   \xmark &  \xmark & 79.30 &85.21   &  86.60   &   86.78   &   86.43 & 84.86   \\ 
                                           &  & \cmark & \xmark & 84.00 &85.73  &  85.73         &  86.43   &  86.08& 85.60\\ 
                                           &  & \xmark  & \cmark & 83.13 & 85.39 &  85.73       & 86.78   &  86.78& 85.56 \\
                                           &  & \cmark  & \cmark &  85.21& 85.39 & 85.73 &    \bf87.13  &  \bf 87.30 & \bf 86.15  \\
               Avg.                     &  &    -     &  -     &  82.91           &   85.43   &  85.95   &     \bf86.78     &  \bf86.65          &  -\\ 
\hline 
  \end{tabular}}
\end{center}
\caption{Ablation study on SBU and FPHA databases, when symmetry (sym) and orthogonality (orth) are taken separately and when combined (in these results $K=4$). Avg stands for average performances.}\label{table23}
\end{table}

   \begin{figure*}[tbp]
\center
\includegraphics[width=0.3\linewidth]{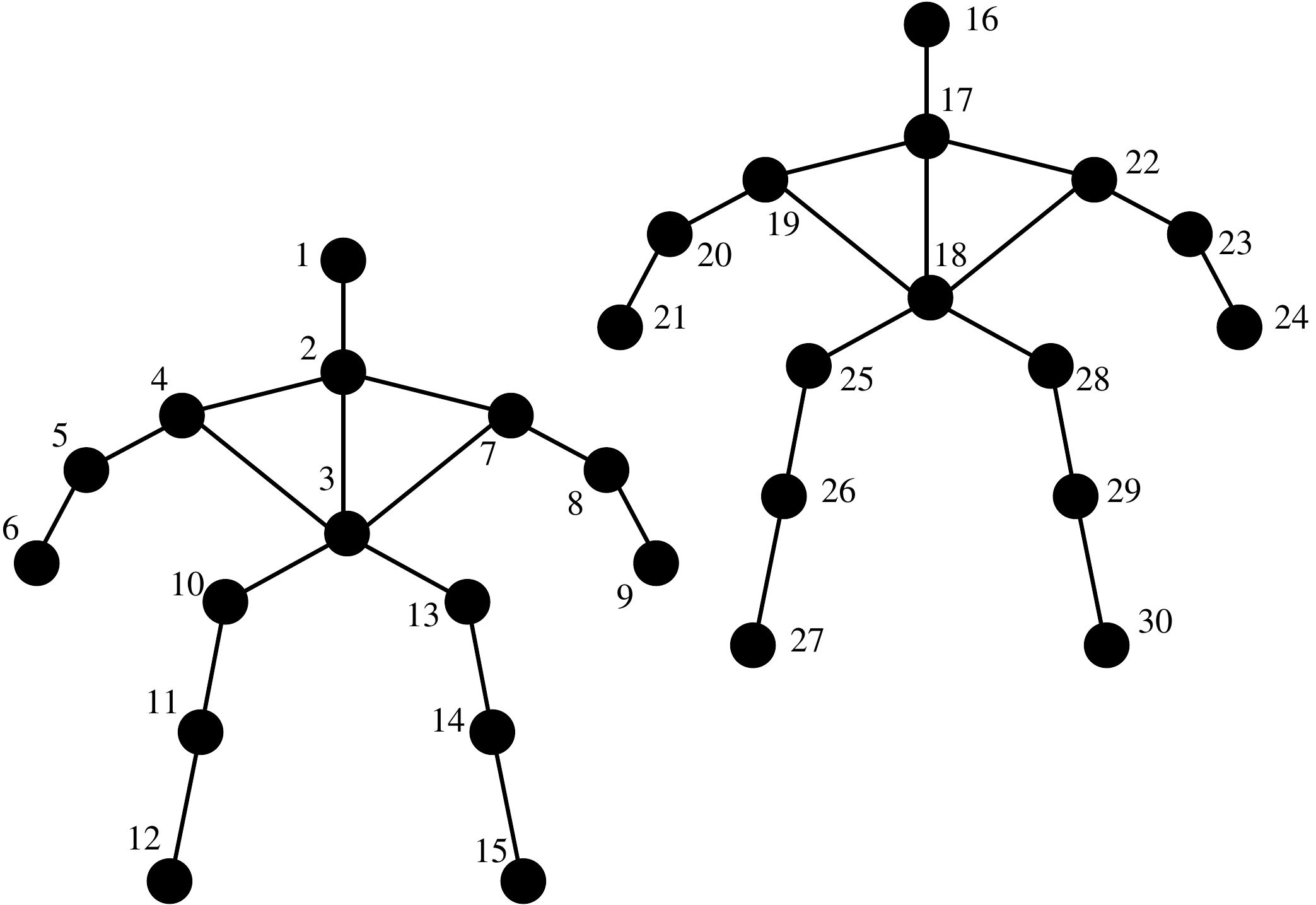} \includegraphics[width=0.3\linewidth]{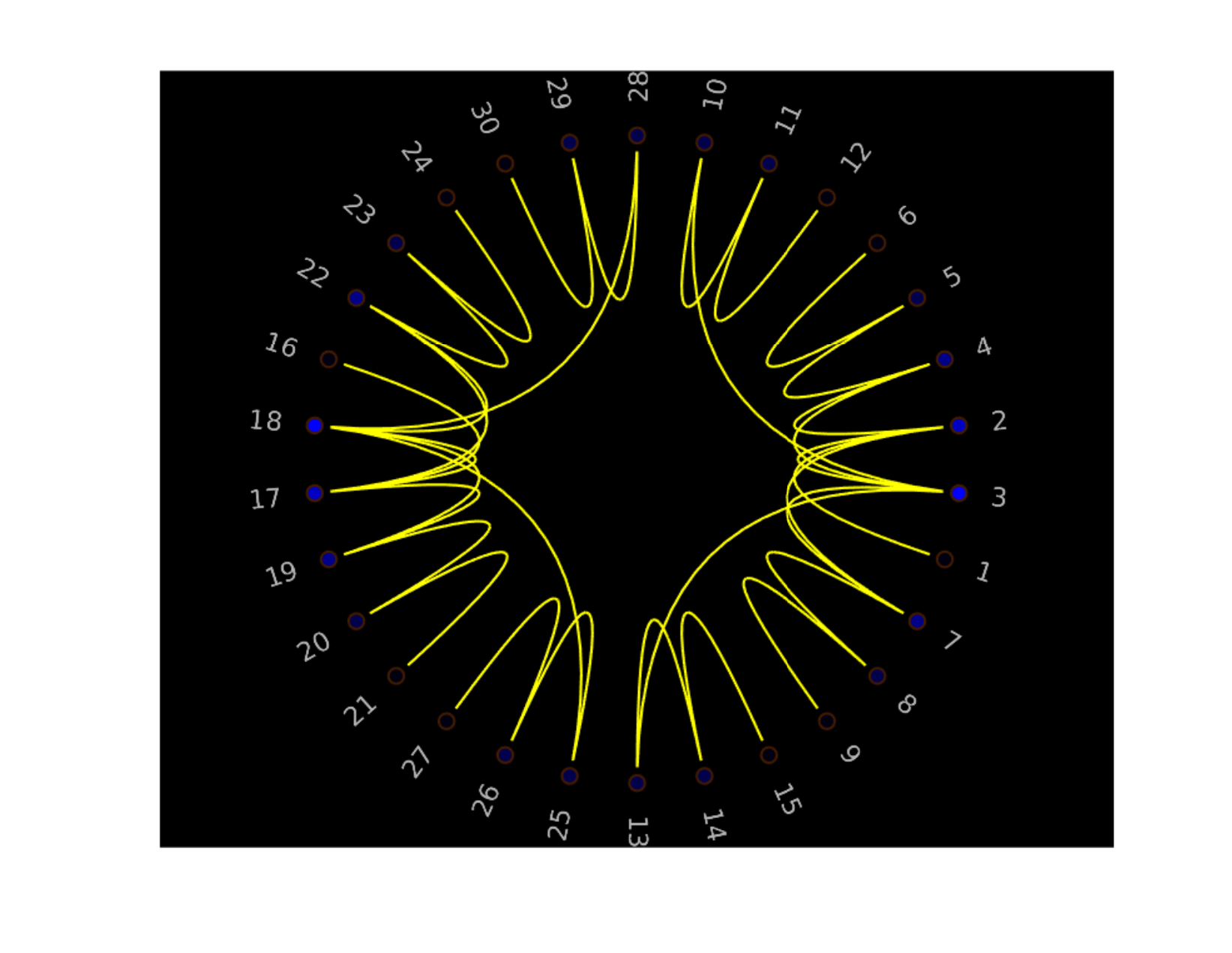} \includegraphics[width=0.3\linewidth]{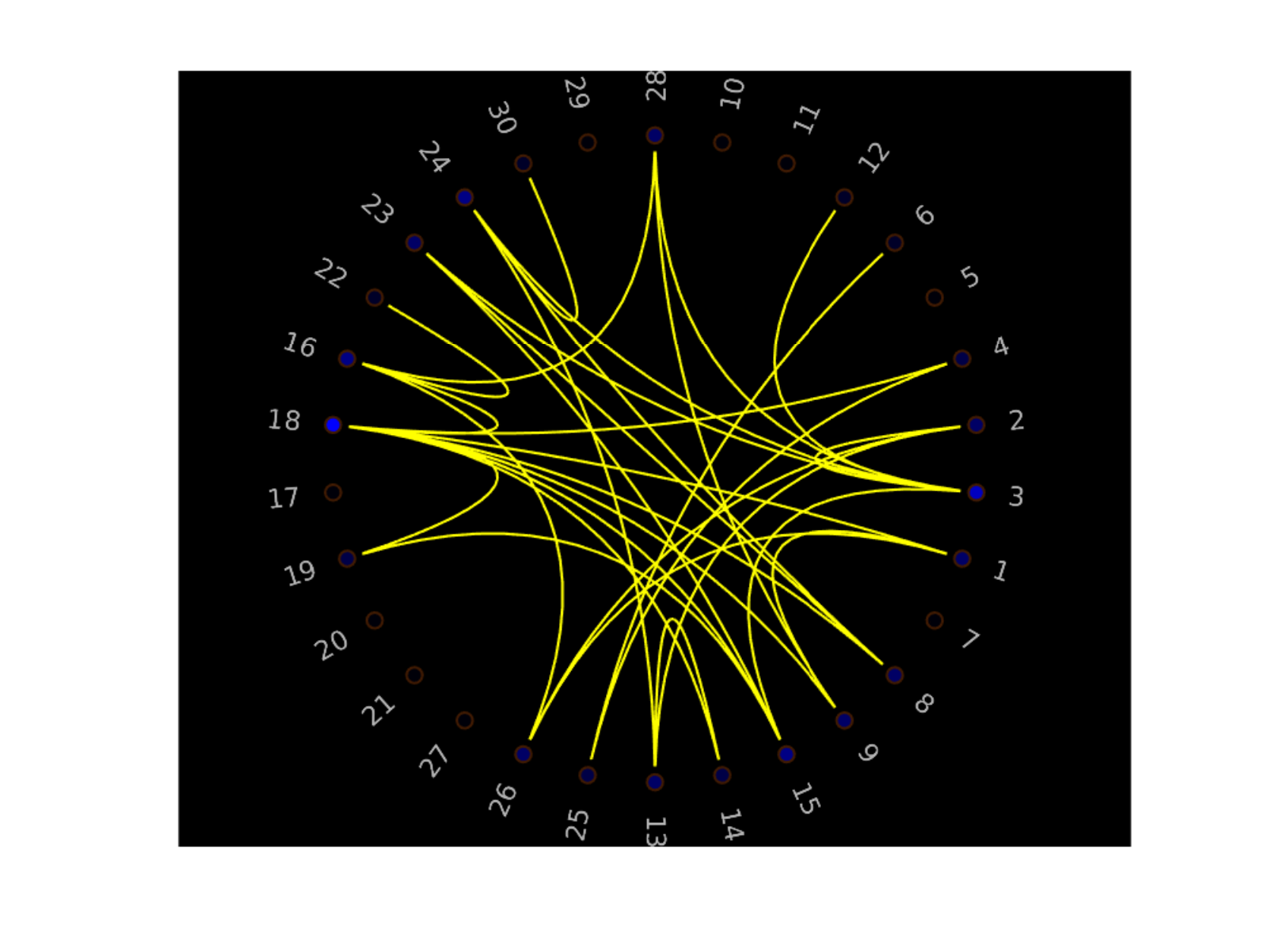} \\
\includegraphics[width=0.15\linewidth]{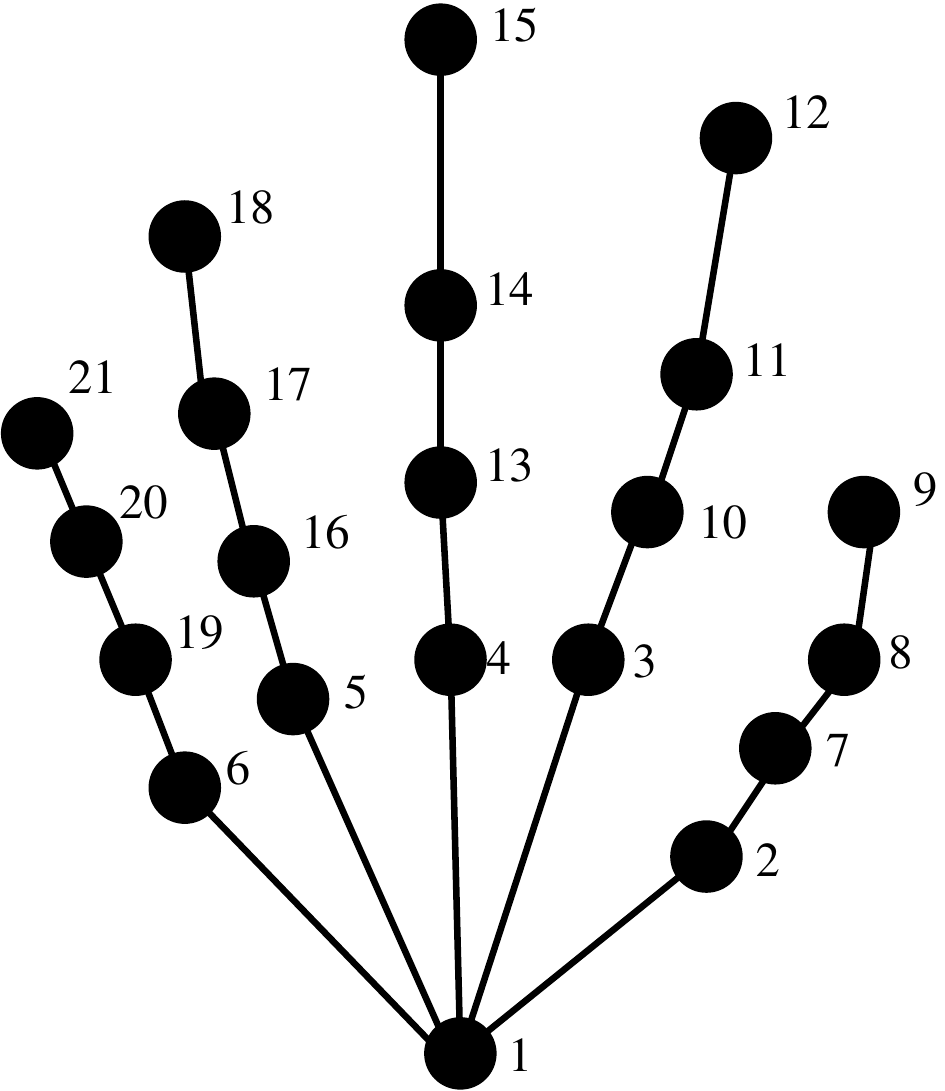} \hspace{2.6cm}\includegraphics[width=0.3\linewidth]{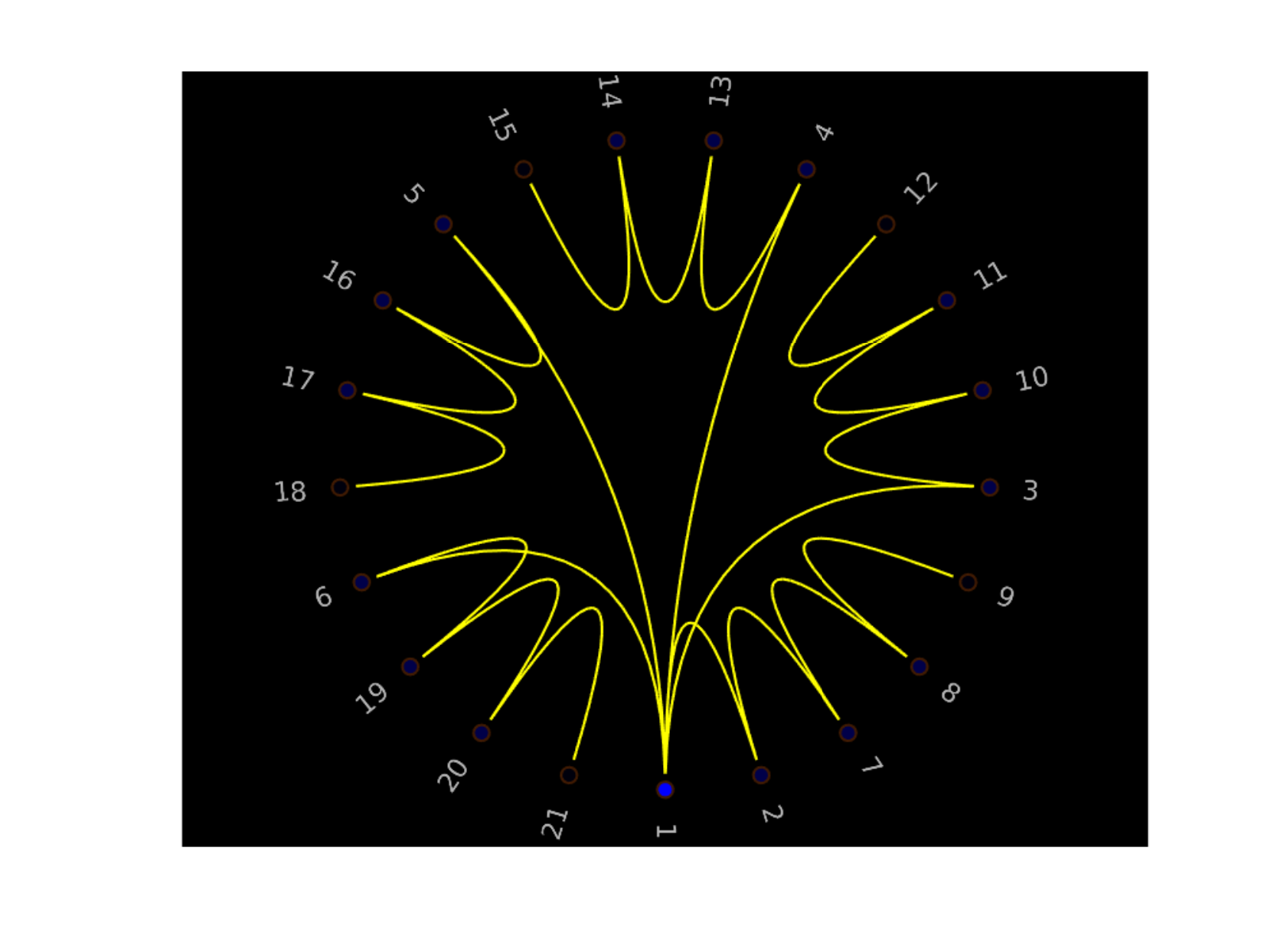} \includegraphics[width=0.3\linewidth]{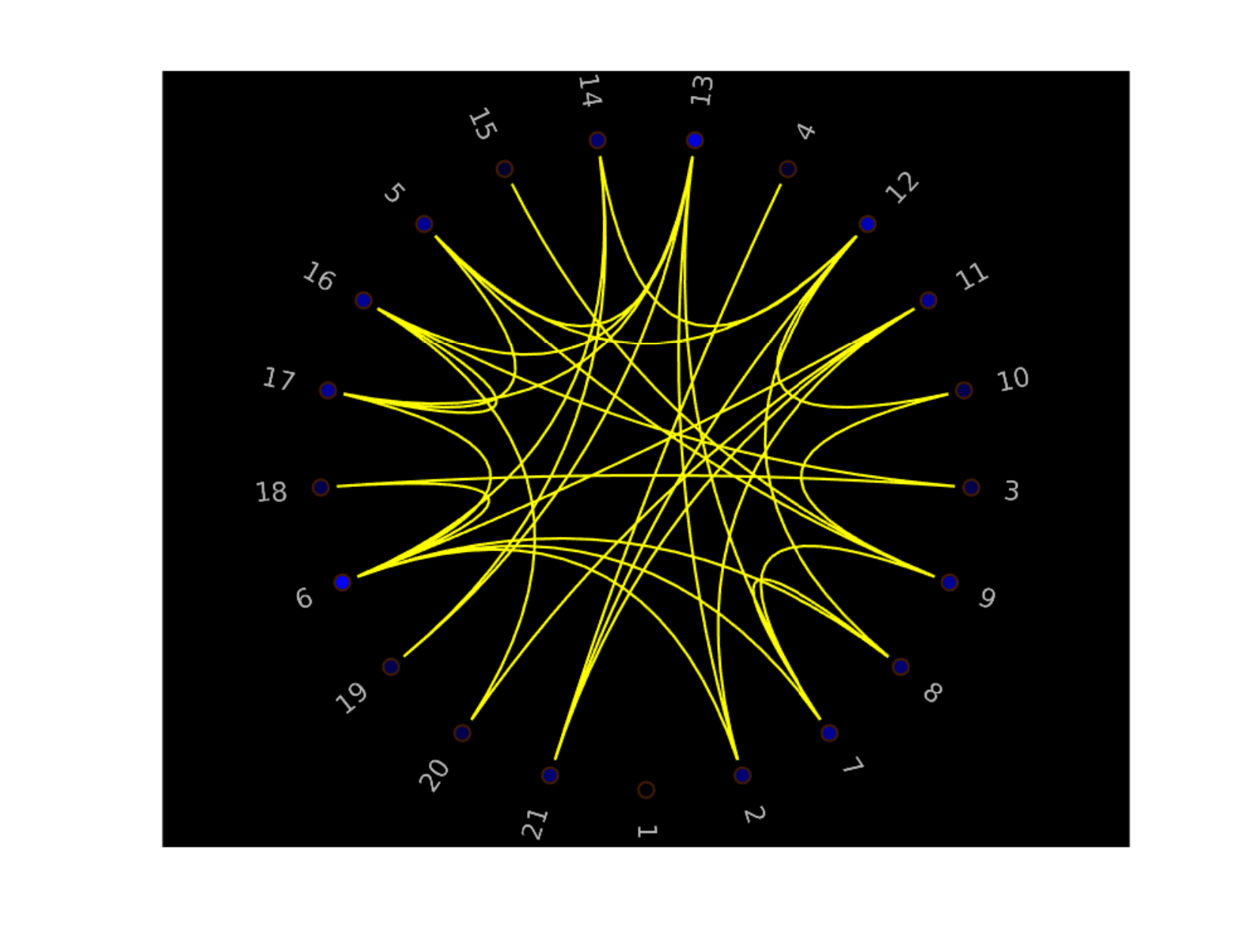} 
\caption{This figure shows original skeletons (left) with their intrinsic node-to-node relationships useful for {\it individual identification} (middle), and an example of the adjacency matrix associated to the learned Laplacian which shows the extrinsic node-to-node relationships found to be the most discriminating for {\it skeleton-based action recognition} when using our proposed  method (the exact setting corresponds to Tables~\ref{table21} and \ref{table22}, using NDRW for SBU and DN for FPHA  both with $K=4$). {\bf (Better to zoom the PDF version to view the learned node-to-node relationships).}}
\label{fig:A2}
\end{figure*}

\begin{table}[ht]
\begin{center}
\resizebox{0.59\columnwidth}{!}{
\begin{tabular}{cc|c}
{\bf Method}      &   & {\bf Accuracy (\%)}\\
\hline 
  Raw Position \cite{Yun2012} & CVPRW 2012  & 49.7   \\ 
  Joint feature \cite{Ji2014}  & ICMEW 2014  & 86.9   \\
  CHARM \cite{Li2015a}       & ICCV2015    & 86.9   \\
 \hline  
H-RNN \cite{Du2015}         & CVPR 2015   & 80.4   \\ 
ST-LSTM \cite{Liu2016}      & ECCV 2016   & 88.6    \\ 
Co-occurrence-LSTM \cite{Zhua2016} & AAAI 2016 & 90.4  \\ 
STA-LSTM  \cite{Song2017}     & AAAI 2017   & 91.5  \\ 
ST-LSTM + Trust Gate \cite{Liu2016} & ECCV 2016 & 93.3 \\
VA-LSTM \cite{Zhang2017}      & ICCV 2017 & 97.6  \\
 GCA-LSTM \cite{GCALSTM}                    &    TIP 2018     &  94.9     \\ 
  \hline
Riemannian manifold. traj~\cite{RiemannianManifoldTraject} &  PAMI 2018 & 93.7 \\
DeepGRU  \cite{DeepGRU}        &    ISVC 2019  &    95.7    \\
RHCN + ACSC + STUFE \cite{Jiang2020} & WACV 2020  & 98.7 \\ 
  \hline
\hline 
  Multi-Laplacians (ML baseline)  &              &        98.4      \\
  Our  best (table~\ref{table21})              &              &        \bf100                                           
\end{tabular}}
 \end{center} 
  \caption{Comparison against  state of the art methods using the SBU database.}\label{compare}
 \end{table} 

\begin{table}[ht]
\begin{center}
\resizebox{0.59\columnwidth}{!}{
\begin{tabular}{cccc|c}
{\bf Method} & {\bf Color} & {\bf Depth} & {\bf Pose} & {\bf Accuracy (\%)}\\
\hline
  Two stream-color \cite{refref10}   & \cmark  &  \xmark  & \xmark  &  61.56 \\
Two stream-flow \cite{refref10}     & \cmark  &  \xmark  & \xmark  &  69.91 \\ 
Two stream-all \cite{refref10}      & \cmark  & \xmark   & \xmark  &  75.30 \\
\hline 
HOG2-depth \cite{refref39}        & \xmark  & \cmark   & \xmark  &  59.83 \\    
HOG2-depth+pose \cite{refref39}   & \xmark  & \cmark   & \cmark  &  66.78 \\ 
HON4D \cite{refref40}               & \xmark  & \cmark   & \xmark  &  70.61 \\ 
Novel View \cite{refref41}          & \xmark  & \cmark   & \xmark  &  69.21  \\ 
\hline
1-layer LSTM \cite{Zhua2016}        & \xmark  & \xmark   & \cmark  &  78.73 \\
2-layer LSTM \cite{Zhua2016}        & \xmark  & \xmark   & \cmark  &  80.14 \\ 
\hline 
Moving Pose \cite{refref59}         & \xmark  & \xmark   & \cmark  &  56.34 \\ 
Lie Group \cite{Vemulapalli2014}    & \xmark  & \xmark   & \cmark  &  82.69 \\ 
HBRNN \cite{Du2015}                & \xmark  & \xmark   & \cmark  &  77.40 \\ 
Gram Matrix \cite{refref61}         & \xmark  & \xmark   & \cmark  &  85.39 \\ 
TF    \cite{refref11}               & \xmark  & \xmark   & \cmark  &  80.69 \\  
\hline 
JOULE-color \cite{refref18}         & \cmark  & \xmark   & \xmark  &  66.78 \\ 
JOULE-depth \cite{refref18}         & \xmark  & \cmark   & \xmark  &  60.17 \\ 
JOULE-pose \cite{refref18}         & \xmark  & \xmark   & \cmark  &  74.60 \\ 
JOULE-all \cite{refref18}           & \cmark  & \cmark   & \cmark  &  78.78 \\ 
\hline 
Huang et al. \cite{Huangcc2017}     & \xmark  & \xmark   & \cmark  &  84.35 \\ 
Huang et al. \cite{ref23}           & \xmark  & \xmark   & \cmark  &  77.57 \\  
  \hline
Our best (table~\ref{table22})                    & \xmark  & \xmark   & \cmark  & \bf87.3                                                 
\end{tabular}}
\end{center} 
  \caption{Comparison against  state of the art methods using the FPHA database.}\label{compare2}
 \end{table} 
 
\section{Conclusion} 
In this paper, we introduce a novel Chebyshev-based Laplacian design for graph convolutional networks (GCNs). The learned Laplacian operators capture the most influencing interactions between body parts in skeleton-based action recognition. The strength of our method resides in its ability to learn shared Laplacians which are embedded into a Chebyshev polynomial basis that allows increasing the discrimination power of our graph representations. In contrast to many existing networks, the parameters of our GCN are interpretable as their design is constrained ``by construction'' and this also acts as  a regularizer that mitigates overfitting. Indeed, our  proposed   parametrization ---  together with Laplacian weight sharing,  symmetry and orthogonality  ---  enhance the representational  power of our learned GCNs without increasing their actual number of the training parameters.  Several Laplacian  operators are also considered including differential and non-differential ones which model the statistical properties of the learned graph representations, and when combined, they further enhance the performances of action recognition.  Extensive experiments, conducted on standard databases (namely SBU and FPHA) show a clear gain of our design w.r.t. different handcrafted and (other) learned Laplacians, as well as the related work in skeleton-based action recognition. \\
As a future work, we are currently investigating the extension of our method to other applications relying on symmetric positive definite (SPD)  matrices in order to aggregate the convolutional features on SPD manifolds \cite{Harandia2018,Huanga2018,Huangb2015,Zhanggg2017}.

\end{document}